\documentclass[runningheads]{llncs}
\usepackage{eccv}
\usepackage{eccvabbrv}
\usepackage{graphicx}
\usepackage{booktabs}
\usepackage{amsmath}
\usepackage{amssymb}
\usepackage[accsupp]{axessibility}
\usepackage[pagebackref,breaklinks,colorlinks,citecolor=eccvblue]{hyperref}
\usepackage{orcidlink}
\usepackage{xcolor}

\begin{document}

\title{Conserved Immune Topology Improves Pathology Foundation Model Generalization for Cross-Cancer MSI-H Prediction\thanks{\textcolor{blue}{\textbf{Accepted at the ECCV 2026 Workshop on Medical Foundation Models and Benchmarks (MedFM-Bench).}}}}

\titlerunning{CIT for Cross-Cancer MSI Prediction}
\author{Dasari Naga Raju}
\authorrunning{Dasari Naga Raju}
\institute{Independent Researcher, Andhra Pradesh, India \\
\textsuperscript{} Corresponding Author: \email{raajuuu1998@gmail.com}}
\maketitle

\begin{abstract}
Pathology foundation models integrated with multiple instance learning achieve competitive accuracy within single-cancer cohorts, yet cross-cancer generalization remains unresolved due to organ-specific histological and architectural differences. In this paper, we propose Conserved Immune Topology (CIT), a lightweight spatial representation for cross-cancer MSI-H prediction that augments foundation-model embeddings with biologically motivated immune descriptors. CIT uses unsupervised clustering to identify immune-associated tiles, then encodes tertiary lymphoid structures, peritumoral immune reactions, multi-scale tumor-infiltrating lymphocyte density, and immune--tumor mixing from frozen foundation-model embeddings and tile coordinates without requiring annotations or target-domain data. The proposed method was evaluated under cross-site and cross-cancer settings using CPTAC-COAD and TCGA-STAD cohorts, which introduce scanner variability, distribution shifts, and organ-specific architectural variations. Zero-shot cross-cancer transfer with CIT increased TransMIL AUC from 0.6627 to 0.7161, an absolute gain of 0.0534 ($p=0.003$), with consistent improvements across all three MIL aggregators. These results suggest that spatial immune topology provides potentially an organ-invariant representation for MSI-H prediction, supporting cross-cancer generalization of pathology foundation models.

\end{abstract}

\section{Introduction}
\label{sec:intro}
Microsatellite instability-high (MSI-H) is an important biomarker for prognosis, treatment selection, and immunotherapy decisions across multiple cancer types~\cite{ref-boland2010,ref-le2017}. MSI-H status is conventionally determined using molecular assays and immunohistochemistry-based evaluation of mismatch repair proteins. These approaches require additional laboratory tests, motivating the development of computational methods that infer the status of MSI-H directly from whole-slide images. Recent studies in computational pathology have applied multiple instance learning (MIL) and pathology foundation models to MSI-H predictions. Foundation models have demonstrated strong performance in diagnostic and prognostic tasks and, when combined with MIL, achieve competitive performance in MSI-H prediction within individual types of cancer.

However, cross-cancer transfer introduces domain gaps due to organ-specific histological differences. Most existing methods have been developed and evaluated within a single cancer cohort, where training and testing samples share similar tissue characteristics. Applying models across cancer types introduces distribution shifts arising from organ-specific histological and architectural differences. Consequently, representations learned from one cancer type often transfer poorly to another. In contrast, MSI-associated immune phenomena are conserved across organs. MSI-H tumors are characterized by mismatch repair deficiency and demonstrate common immune-related patterns including tertiary lymphoid structures, peritumoral lymphocytic reactions, increased tumor-infiltrating lymphocyte (TIL) density, and immune--tumor mixing, observed across colorectal, gastric, and other cancer types~\cite{ref-smyrk2001,ref-mlecnik2016,ref-saltz2018til}. These observations motivate the hypothesis that spatial immune organization encodes organ-invariant correlates of MSI-H status.

Motivated by this observation, we propose Conserved Immune Topology (CIT), a lightweight spatial representation for cross-cancer MSI-H prediction. CIT augments foundation-model embeddings with biologically motivated spatial descriptors characterizing immune organization. The resulting representation integrates into existing MIL pipelines without requiring annotations, target-domain adaptation, or modifications to the underlying foundation model, demonstrating improved generalization and robustness under distribution shift.

\section{Related Work}
\label{sec:related}

\subsection{MSI Prediction in Computational Pathology}

Whole-slide image (WSI) analysis for MSI-H detection combines morphological pattern recognition with foundation-model embeddings and multiple-instance aggregation. Early work by Kather et al.~\cite{ref-kather2019} demonstrated that mismatch repair deficiency leaves morphological signatures detectable directly from histopathological images. Further studies confirmed these findings across larger multicenter cohorts and reported improved performance through MIL and attention-based aggregation methodologies~\cite{ref-echle2020gastro,ref-yamashita2021}. Subsequent work introduced weakly supervised and self-supervised methods to reduce annotation burden while maintaining performance, including iterative tile-sampling approaches for molecular pathways and mutation prediction~\cite{ref-bilal2021}. Contrastive self-supervised pretraining with heterogeneity-aware aggregation has also been applied to mismatch repair-related biomarkers across colorectal and breast cancer cohorts~\cite{ref-schirris2022}. Recent approaches adopted pathology foundation models as feature extractors, achieving competitive performance for MSI-H prediction within single cancer cohorts~\cite{ref-wagner2023}. However, within-cohort validation typically reports AUCs $\geq0.85$~\cite{ref-kather2019,ref-echle2020}, while cross-cancer transfer studies remain sparse, with reported performance degrading $0.15$--$0.25$ AUC units~\cite{ref-cheung2026}.

Recent pathology foundation models enable extraction of transferable representations from large collections of histopathological images. The pathology foundation models such as UNI~\cite{ref-chen2024}, UNI2-h~\cite{ref-uni2h}, CONCH~\cite{ref-chen2023conch}, and Virchow2~\cite{ref-zimmermann2024virchow2} achieve $>0.90$ AUC on downstream diagnostic tasks including tumor grade, mutational status, and survival prediction~\cite{ref-chen2024}. These representations serve as frozen feature extractors within MIL pipelines, reducing the need for task-specific feature engineering. Despite effectiveness within cohorts, foundation-model representations remain sensitive to domain-specific appearance and acquisition characteristics when applied across datasets or cancer types~\cite{ref-cheung2026}.

MIL has become the standard framework for slide-level prediction from whole-slide histopathological images. Attention-based MIL (ABMIL)~\cite{ref-ilse2018} introduced learnable attention mechanisms to aggregate tile features into slide-level representations. CLAM~\cite{ref-lu2021} extended this formulation through instance-level clustering, whereas TransMIL~\cite{ref-shao2021} incorporated transformer-based interactions between image tiles. These methods differ primarily in their aggregation mechanisms on frozen pathology foundation-model embeddings. Comparative benchmarking studies have systematically evaluated weakly supervised MIL pipelines, highlighting the impact of feature extraction and aggregation strategies on slide-level performance~\cite{ref-laleh2022}.

\subsection{Generalization and Spatial Immune Representations}

Generalization across datasets and institutions remains an important challenge in computational pathology. Existing approaches evaluate stain normalization, data augmentation, domain adaptation, and domain generalization techniques to reduce site-specific variations~\cite{ref-tellez2019,ref-macenko2009}. In biomarker prediction, multi-centric studies have reported that models trained for MSI-H prediction frequently transfer poorly to independent external cohorts~\cite{ref-niehues2023}. Domain adaptation methods require target-domain data during training; conversely, domain generalization techniques learn invariant representations without target-domain supervision~\cite{ref-cheung2026}. While these methods primarily address acquisition and staining differences, cross-cancer transfer introduces additional challenges from organ-specific histology and architecture. These limitations motivate transferable spatial representations for MSI-H prediction across cancer types.

Recent studies characterize the spatial organization of the tumor immune microenvironment beyond tissue appearance. Deep learning has been used to map TILs and relate their spatial distribution to molecular and clinical correlates~\cite{ref-saltz2018til}. Handcrafted spatial descriptors quantifying TIL arrangement and colocalization have been shown to be prognostic of recurrence~\cite{ref-corredor2019}, and dedicated models have been proposed for automated detection of tertiary lymphoid structures~\cite{ref-vanrijthoven2024}. These spatial immune patterns are particularly relevant to MSI-H tumors, which exhibit similar immune organization across multiple cancer types. CIT infers immune organization directly from frozen foundation-model embeddings and tile coordinates without relying on supervised cell- or structure-level annotations.
\section{Method}
\label{sec:method}
The CIT framework constructs spatial immune descriptors from frozen pathology foundation-model embeddings and tile coordinates. CIT characterizes immune-associated regions through unsupervised clustering and computes four distinct groups of spatial descriptors encoding tertiary lymphoid structures, peritumoral immune reactions, multi-scale tumor-infiltrating lymphocyte density, and immune--tumor mixing. The ten-dimensional descriptor vector is concatenated with the corresponding foundation-model embedding to form an augmented tile representation that an MIL model aggregates for slide-level MSI-H prediction.

\subsection{Preliminaries and Notation}
\label{sec:prelim}
A WSI is partitioned into $N$ image tiles, each represented by a frozen pathology foundation model $\phi$ that extracts tile embeddings
\[
\mathbf{F}=\{\mathbf{f}_1,\mathbf{f}_2,\ldots,\mathbf{f}_N\}\in\mathbb{R}^{N\times d},
\]
where $d=1536$ for UNI2-h and $d=512$ for CONCH. Each tile $i$ is associated with spatial coordinate $\mathbf{c}_i=(x_i,y_i)$ within a slide of width $W$ and height $H$. Spatial coordinates are normalized by slide dimensions to enable transfer across images with varying resolutions and fields of view:
\[
\tilde{\mathbf{c}}_i=\left(\frac{x_i}{W},\frac{y_i}{H}\right).
\]
Each embedding is $\ell_2$-normalized, denoted $\hat{\mathbf{f}}_i$. Spatial neighborhoods are defined using Euclidean distance between normalized tile coordinates. We denote by $\mathcal{N}_k(i)$ the set of $k$ nearest spatial neighbors of tile $i$, where $k\in\{10,20,30,100\}$ depends on the specific spatial descriptor. For each tile, CIT computes a ten-dimensional descriptor vector $\mathbf{u}_i\in\mathbb{R}^{10}$, subsequently concatenated with the corresponding foundation-model embedding to form the augmented representation used for slide-level prediction within an MIL framework (\cref{sec:integration}).
\begin{figure}[t]
    \centering
    \includegraphics[width=\textwidth]{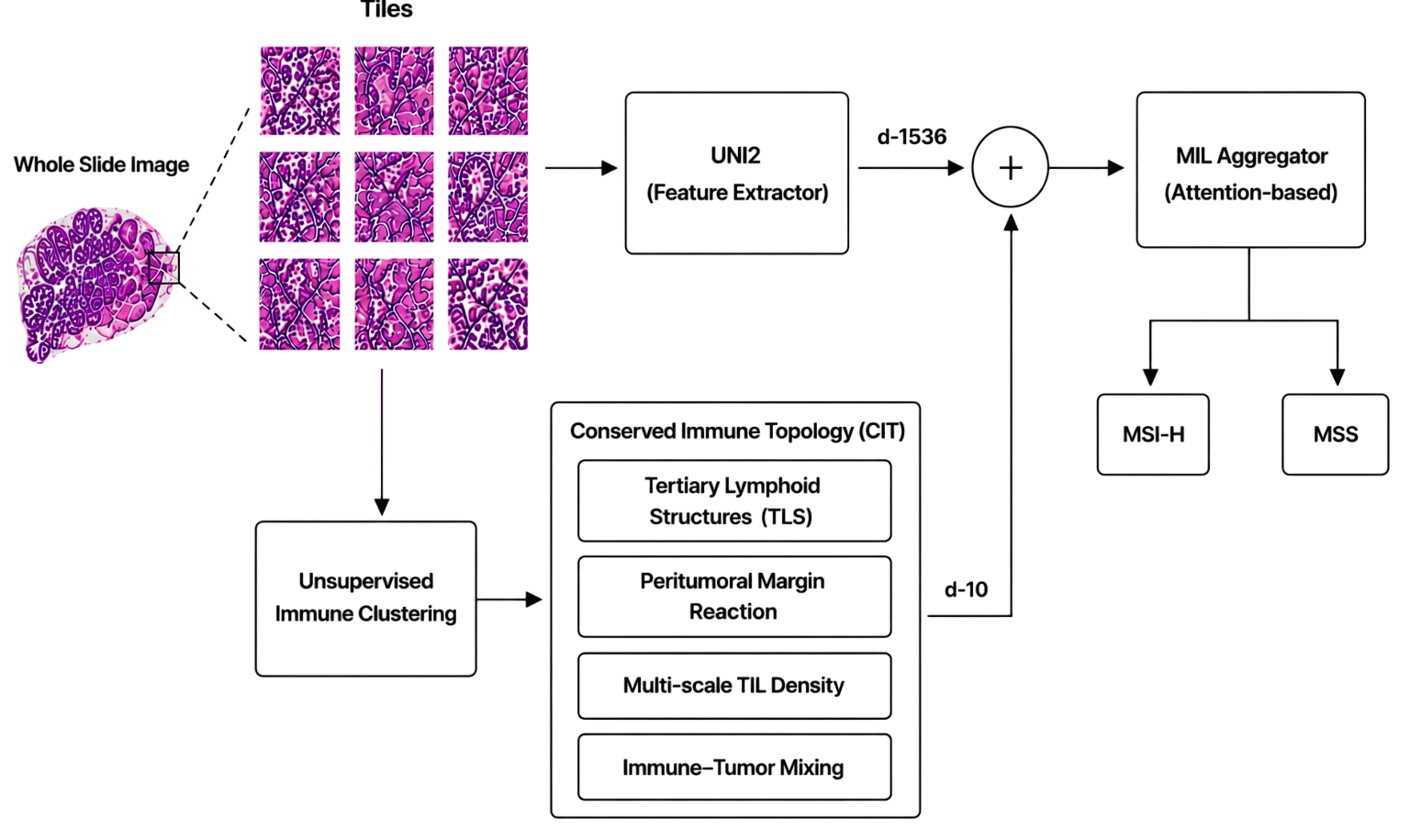}
\caption{Overview of the proposed framework. WSI patches are encoded via UNI2-h and concatenated with a CIT descriptor.}
    \label{fig:framework}
\end{figure}
\subsection{Unsupervised Immune-Tile Identification}
\label{sec:immune}
CIT infers immune-associated regions from frozen foundation-model embeddings using unsupervised clustering without manual annotations or prior knowledge of immune cell morphology. Instead of detecting individual immune cells, the framework infers immune-associated regions by grouping tiles with similar feature representations, which subsequently form the basis for computing spatial descriptors.

K-means clustering ($K=50$, determined via elbow curve on training cohort) was applied to up to $500$ random tiles per slide from the training cohort, balancing computational cost and cluster stability. Because lymphocyte-rich regions exhibit lower morphological heterogeneity, K-means assigns them to tighter clusters in foundation-model space, enabling unsupervised immune enrichment. The cluster tightness score for cluster $c$ is defined as
\begin{equation}
\tau_c=
\left(
\frac{1}{|c|}
\sum_{i\in c}
\|\hat{\mathbf{f}}_i-\boldsymbol{\mu}_c\|_2
+\epsilon
\right)^{-1},
\label{eq:tightness}
\end{equation}
where $\boldsymbol{\mu}_c$ denotes the cluster centroid, $|c|$ is the number of tiles in cluster $c$, $\hat{\mathbf{f}}_i$ is the $\ell_2$-normalized embedding of tile $i$, and $\epsilon$ is a small constant for numerical stability. The tightness score assigns higher values to clusters with lower intracluster variation.

Clusters were ranked by $\tau_c$, and the top $30\%$ ($\kappa=15$ of 50 clusters) by tightness score were designated immune-enriched, a threshold selected using downstream validation performance on held-out TCGA-COAD folds. The identified regions were interpreted as computational proxies for immune-associated tissue rather than explicit cell-type annotations.

Each tile is assigned a binary immune indicator $m_i \in \{0,1\}$ (cluster membership) and continuous immune enrichment score $s_i$, which integrates cluster tightness and membership:
\begin{equation}
s_i=
\frac{\tau_{c(i)}}{\max_j\tau_{c(j)}}(1+m_i),
\label{eq:immunescore}
\end{equation}
where $c(i)$ denotes the cluster assigned to tile $i$, $\tau_{c(i)}$ denotes the corresponding cluster tightness score, and $m_i$ denotes the binary immune indicator. The clustering model and all associated hyperparameters were determined exclusively from the training cohort and remained fixed for all subsequent cross-site and cross-cancer evaluations.

\subsection{Conserved Immune Topology Descriptors}
\label{sec:descriptors}

CIT computes four groups of spatial descriptors encoding conserved immune topology: tertiary lymphoid structures, peritumoral immune reactions, multi-scale immune density, and immune--tumor mixing. Together, these descriptors define a ten-dimensional vector $\mathbf{u}_i$. The overall architecture of CIT is shown in \cref{fig:framework}.

\subsubsection{Tertiary Lymphoid Structures (TLS)}
\label{sec:tls}

Tertiary lymphoid structures (TLS) are organized lymphoid aggregates associated with effective antitumor immune responses in MSI-H tumors~\cite{ref-mlecnik2016}. TLS were detected as compact immune-enriched clusters using DBSCAN ($\varepsilon=0.02$, minimum 5 points; determined via grid search on TCGA-COAD training folds) applied to normalized coordinates of immune-associated tiles ($m_i=1$). Each detected cluster $t$ is represented by its centroid $\mathbf{g}_t$ and cluster size $n_t$. Three TLS descriptors per tile are computed: (1) binary membership to detected TLS clusters, (2) normalized distance to nearest TLS centroid (\cref{eq:tlsdist}), and (3) relative size of nearest cluster. These capture membership, spatial proximity, and TLS size.

The normalized distance from tile to its nearest TLS centroid is:
\begin{equation}
\delta_i^{\mathrm{TLS}}=
\frac{\min_t\|\tilde{\mathbf{c}}_i-\mathbf{g}_t\|_2}
{\max_j\min_t\|\tilde{\mathbf{c}}_j-\mathbf{g}_t\|_2},
\label{eq:tlsdist}
\end{equation}
where $\tilde{\mathbf{c}}_i$ denotes the normalized coordinate of tile $i$, and $\mathbf{g}_t$ is the centroid of TLS cluster $t$.

\subsubsection{Peritumoral Margin Reaction}
\label{sec:margin}

Crohn's-like lymphocytic reaction at the invasive tumor margin is a characteristic histological feature of MSI-H tumors~\cite{ref-jass2007,ref-smyrk2001}. CIT estimates the relative spatial position of each tile from local spatial density without explicit tissue annotations. Because interior regions have denser spatial neighborhoods, local density inversely correlates with margin distance, enabling boundary detection without tissue segmentation. The local spatial density is estimated from the average distance to the $20$ nearest spatial neighbors:
\begin{equation}
\rho_i=
\left(
\frac{1}{20}
\sum_{j\in\mathcal{N}_{20}(i)}
\|\tilde{\mathbf{c}}_i-\tilde{\mathbf{c}}_j\|_2
+\epsilon
\right)^{-1},
\quad
\delta_i^{\mathrm{marg}} =
1- \frac{\rho_i}{\max_j\rho_j},
\label{eq:margin}
\end{equation}
where $\rho_i$ denotes the local spatial density of tile $i$, $\mathcal{N}_{20}(i)$ represents its 20 nearest spatial neighbors, $\delta_i^{\mathrm{marg}}$ is the normalized margin distance, and $\epsilon$ is a small constant.

The peritumoral immune band descriptor combines normalized margin distance with the immune score:
\begin{equation}
b_i=
\frac{s_i\,\delta_i^{\mathrm{marg}}}{\max_j s_j\,\delta_j^{\mathrm{marg}}},
\label{eq:band}
\end{equation}
where $s_i$ denotes the immune score of tile $i$. Higher values of $b_i$ indicate immune-associated regions near the tissue margins.

\subsubsection{Multi-scale Immune Density}
\label{sec:til}

MSI-H tumors are characterized by high densities of TILs, reflecting active antitumor immune response~\cite{ref-saltz2018til}. CIT quantifies the local density of immune-associated tiles across multiple spatial neighborhoods. For each tile, immune density is computed as the fraction of immune-associated tiles within its $k$-nearest-neighbor neighborhood:
\begin{equation}
\pi_i^{(k)} =
\frac{1}{k}
\sum_{j\in\mathcal{N}_k(i)}
m_j,
\label{eq:til}
\end{equation}
where $\pi_i^{(k)}$ denotes the immune density around tile $i$, $\mathcal{N}_k(i)$ represents its $k$ nearest spatial neighbors, $m_j\in\{0,1\}$ is the immune indicator of neighboring tile $j$, and $k\in\{10,30,100\}$. The three neighborhood sizes capture complementary spatial scales from local to regional immune density.

\subsubsection{Immune--Tumor Mixing}
\label{sec:mix}
The spatial arrangement of immune-associated and non-immune regions provides complementary information regarding local immune density. Spatial intermixing of immune and tumor regions is characteristic of effective antitumor immune responses, whereas spatial separation is characteristic of immune exclusion. CIT computes two descriptors quantifying neighborhood heterogeneity and immune-associated tile abundance. Local immune--tumor mixing is quantified by binary entropy of the immune fraction within the 20-nearest-neighbor neighborhood:
\begin{equation}
p_i=
\frac{1}{20}
\sum_{j\in\mathcal{N}_{20}(i)}
m_j,
\label{eq:pi}
\end{equation}
where $p_i$ is the local fraction of immune-associated tiles. The corresponding mixing entropy is:
\begin{equation}
e_i =
-p_i\log_2 p_i
-(1-p_i)\log_2(1-p_i),
\label{eq:entropy}
\end{equation}
with the convention $0\log_2 0=0$. Higher entropy indicates greater immune--tumor mixing, whereas lower values correspond to spatially homogeneous neighborhoods. CIT further computes an immune--tumor ratio over the 30-nearest-neighbor neighborhood:
\begin{equation}
r_i=
\frac{n_i^{\mathrm{imm}}}{30-n_i^{\mathrm{imm}}+1},
\qquad
n_i^{\mathrm{imm}} =
\sum_{j\in\mathcal{N}_{30}(i)}
m_j,
\label{eq:ratio}
\end{equation}
where $n_i^{\mathrm{imm}}$ denotes the number of immune-associated neighboring tiles, and $r_i$ is further normalized across the slide. Entropy measures the degree of local spatial mixing, while the immune--tumor ratio summarizes the relative prevalence of immune-associated tissue within the surrounding neighborhood.

\subsection{Integration with Multiple Instance Learning}
\label{sec:integration}

For each tile, the descriptor vector $\mathbf{u}_i$ is concatenated with the corresponding frozen foundation-model embedding to obtain the augmented representation:
\begin{equation}
\mathbf{f}_i' =
[\mathbf{f}_i \,\Vert\, \mathbf{u}_i] \in \mathbb{R}^{d+10},
\label{eq:concat_final}
\end{equation}
which serves as input to the MIL aggregator. Concatenation was chosen over multimodal fusion because (1) it preserves the frozen foundation model independently, (2) it requires no additional trainable parameters for integration, and (3) the modest 10-dimensional augmentation avoids significant computational overhead. Because CIT operates on frozen tile-level feature representations, it is independent of the aggregation method and integrates into existing MIL frameworks without modifying the feature extractor or aggregation architecture. We evaluated CIT using three representative MIL aggregators: ABMIL, CLAM-SB, and TransMIL.

\section{Experimental Setup}
\label{sec:setup}

\paragraph{Datasets and protocol.}
We evaluated CIT on three publicly available cohorts with molecularly determined MSI labels obtained from cBioPortal~\cite{ref-cerami2012,ref-gao2013}. TCGA-COAD (N=301, 17.9\% MSI-H) was used for training and internal five-fold stratified cross-validation. CPTAC-COAD~\cite{ref-vasaikar2019} (N=105, 22.9\% MSI-H, different institution and scanner) assessed cross-site robustness while preserving the colorectal cancer prediction task. TCGA-STAD (N=308, 17.5\% MSI-H, gastric cancer) evaluated zero-shot cross-cancer transfer without retraining or domain adaptation. Class distribution is summarized in \cref{tab:cohorts}.

\paragraph{Feature extraction.}
We extracted $224\times224$ image tiles at $20\times$ magnification ($0.5\,\mu$m/pixel) from tissue regions identified using Otsu thresholding. Tile coordinates were retained for subsequent spatial descriptor computation. UNI2-h~\cite{ref-uni2h} was used as the primary pathology foundation model, with tile features extracted using the frozen model yielding 1536-dimensional embeddings. Internal evaluation included CONCH~\cite{ref-chen2023conch} (512-dimensional embeddings) to compare the framework across pathology foundation models prior to cross-site and cross-cancer experiments.

\paragraph{Aggregators and configurations.}
We evaluated ABMIL~\cite{ref-ilse2018}, CLAM-SB~\cite{ref-lu2021}, and TransMIL~\cite{ref-shao2021} in two configurations: baseline using foundation-model embeddings and CIT variant using augmented tile representations. Architecture and hyperparameters were kept identical across configurations. Internal validation included experiments with both UNI2-h and CONCH, whereas all cross-site and cross-cancer experiments used UNI2-h.

\paragraph{Training and statistics.}
Proposed and baseline models were trained for 40 epochs using the Adam optimizer~\cite{ref-adam} with learning rate $10^{-4}$, weight decay $10^{-5}$, and hidden dimension 256. For each fold, the checkpoint with the highest validation AUC was retained for evaluation. The clustering model from \cref{sec:immune} was learned from training data and applied without modification during cross-site and cross-cancer evaluations. For external evaluation, predictions were averaged across the five cross-validation folds. Statistical significance between baseline and CIT models was assessed using the paired DeLong test for correlated AUCs~\cite{ref-delong1988}. For the primary cross-cancer transfer comparison, we report Bonferroni-adjusted 98.33\% DeLong confidence intervals ($\alpha=0.0167$ for three comparisons).

\begin{table}[tb]
\centering
\caption{Patient counts and MSI-H/MSS class distribution for study cohorts.}
\label{tab:cohorts}
\small
\begin{tabular}{lcccc}
\toprule
Dataset & Patients & MSI-H & MSS & MSI-H (\%) \\
\midrule
TCGA-COAD  & 301 & 54 & 247 & 17.9 \\
TCGA-STAD  & 308 & 54 & 254 & 17.5 \\
CPTAC-COAD & 105 & 24 & 81  & 22.9 \\
\bottomrule
\end{tabular}
\end{table}

\section{Results}
\label{sec:results}

\subsection{Within-Cancer Validation: Internal and Cross-Site Performance}
\label{sec:res-internal}

\Cref{tab:internal} presents internal TCGA-COAD performance for both foundation models across all three MIL aggregators. UNI2-h consistently outperformed CONCH across all aggregators (mean $\Delta$AUC $+0.02$--$0.03$), with lower fold-wise variance ($\pm0.02$ vs. $\pm0.05$), and was selected for subsequent evaluation. Internal validation (TCGA-COAD) showed minimal CIT benefit: ABMIL +0.001 AUC, CLAM-SB $-0.002$ AUC, TransMIL +0.017 AUC, consistent with a ceiling effect on within-cohort tasks. These changes fall within fold-wise variation (SD $\pm0.01$--$0.06$), confirming CIT augmentation maintains within-cohort accuracy.
\begin{table}[!tbp]
\caption{\small Internal TCGA-COAD performance ($301$ patients, $54$ MSI-H / $247$ MSS; MSI-H AUC, mean $\pm$ SD over five folds).}
\label{tab:internal}
\centering
\small
\begin{tabular}{lc}
\toprule
\textbf{Model} & \textbf{MSI-H AUC} \\
\midrule
CONCH + ABMIL             & $0.9145 \pm 0.0499$ \\
CONCH + CLAM-SB           & $0.9181 \pm 0.0563$ \\
CONCH + TransMIL          & $0.9198 \pm 0.0615$ \\
\midrule
UNI2-h + ABMIL            & $0.9337 \pm 0.0166$ \\
UNI2-h + ABMIL + CIT      & $0.9348 \pm 0.0197$ \\
UNI2-h + CLAM-SB          & $0.9469 \pm 0.0233$ \\
UNI2-h + CLAM-SB + CIT    & $0.9447 \pm 0.0205$ \\
UNI2-h + TransMIL         & $0.9398 \pm 0.0274$ \\
UNI2-h + TransMIL + CIT   & $\mathbf{0.9567 \pm 0.0112}$ \\
\bottomrule
\end{tabular}
\end{table}

After validating CIT's within-cohort performance, we next evaluated cross-site generalization under distribution shift. We evaluated models trained on TCGA-COAD using the independent CPTAC-COAD cohort to assess cross-site generalization. CPTAC-COAD comprises colorectal WSIs collected from a different institution using different whole-slide scanners, introducing variations in patient population, acquisition protocols, and imaging characteristics while preserving the colorectal cancer prediction task. \Cref{tab:cptac} summarizes cross-site evaluation results. CIT increased CLAM-SB from 0.7989 to 0.8344 AUC and TransMIL from 0.7870 to 0.8277 AUC on independent CPTAC-COAD. ABMIL showed comparable performance with and without CIT (0.8089 vs. 0.8086 AUC). Cross-site gains ranged from $-0.0003$ (ABMIL) to $+0.0407$ (TransMIL) AUC, with CLAM-SB and TransMIL improving under distribution shift, whereas ABMIL was unchanged.

\begin{table}[!tbp]
\caption{Cross-site generalization on the independent CPTAC-COAD cohort. CIT improves CLAM-SB and TransMIL despite institution and scanner shift, while ABMIL is unchanged.}
\label{tab:cptac}
\centering
\small
\setlength{\tabcolsep}{6pt}
\begin{tabular}{lcc}
\toprule
\textbf{Model} & \textbf{Baseline} & \textbf{CIT} \\
\midrule
ABMIL    & $\mathbf{0.8089}$ & 0.8086 \\
CLAM-SB  & 0.7989 & $\mathbf{0.8344}$ \\
TransMIL & 0.7870 & $\mathbf{0.8277}$ \\
\bottomrule
\end{tabular}
\end{table}
\subsection{Zero-Shot Cross-Cancer Generalization}
\label{sec:res-transfer}

We evaluated the proposed CIT framework under zero-shot cross-cancer transfer, the primary focus of this study. \Cref{tab:transfer} summarizes performance of baseline and CIT-augmented models on independent TCGA-STAD cohort across all three MIL aggregators.

\begin{figure}[tb]
  \centering
\includegraphics[width=0.5\linewidth]{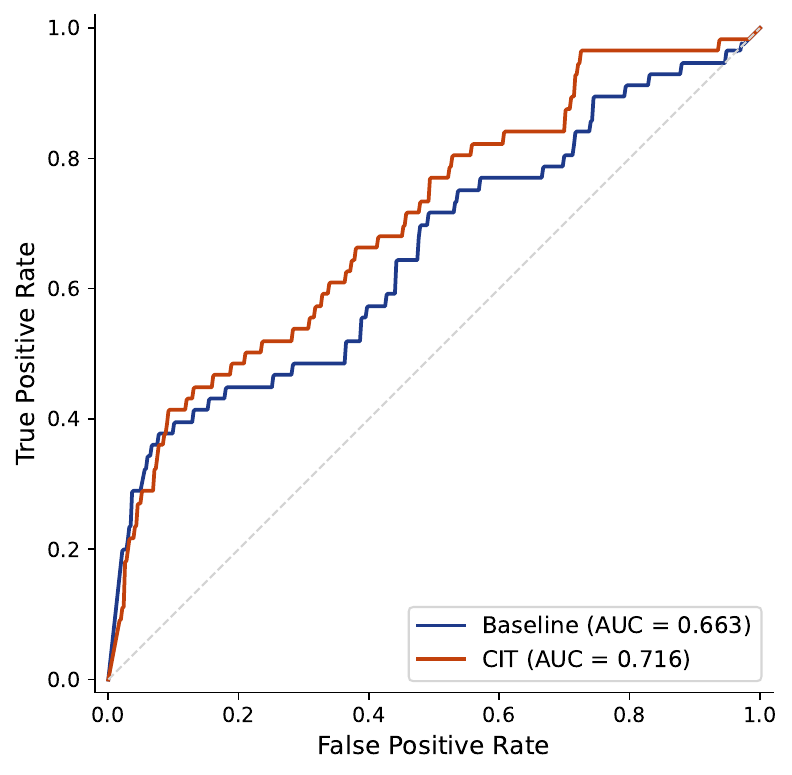}
  \caption{ROC curves for TransMIL with and without CIT on TCGA-STAD. CIT improves the true positive rate across most operating points.}
  \label{fig:roc}
\end{figure}

CIT increased AUC across all three MIL aggregators, with gains of 0.0632, 0.0172, and 0.0534 for ABMIL, CLAM-SB, and TransMIL, respectively. After Bonferroni correction for multiple comparisons ($\alpha=0.0167$), the performance improvements achieved by ABMIL ($p<0.001$) and TransMIL ($p=0.003$) remained statistically significant. In contrast, the improvement observed for CLAM-SB was not statistically significant ($p=0.037$), as reflected by its 98.33\% confidence interval spanning zero. TransMIL improved from 0.6627 to 0.7161 AUC (\cref{fig:roc}), achieving an absolute gain of 0.0534 under zero-shot cross-cancer transfer. For reference, within-cohort MSI prediction on gastric cohorts achieves $\approx0.80$ AUC~\cite{ref-kather2019,ref-echle2020}; zero-shot cross-cancer transfer without spatial features degrades to 0.6627 AUC, whereas CIT recovers 0.7161, narrowing the gap toward within-cohort performance. Across all architectures, the baseline models exhibited limited cross-cancer transfer performance, with AUCs ranging from 0.5681 to 0.6627 on TCGA-STAD. Incorporating CIT shifted the performance range to 0.6313 to 0.7161 AUC, with improvements observed consistently for every MIL aggregator.
\begin{table}[!tbp]
\caption{\small Zero-shot cross-cancer evaluation on TCGA-STAD ($308$ patients; $54$ MSI-H / $254$ MSS), models trained on TCGA-COAD. $\Delta$AUC with 98.33\% DeLong CIs (Bonferroni-adjusted, $\alpha=0.0167$).}
\label{tab:transfer}
\centering
\small
\setlength{\tabcolsep}{6pt}
\begin{tabular}{lcccc}
\toprule
\textbf{Model} & \textbf{Baseline} & \textbf{CIT} &
\textbf{$\Delta$AUC (98.33\% CI)} & \textbf{DeLong $p$} \\
\midrule
ABMIL    & 0.5681 & $\mathbf{0.6313}$ & $+0.0632$~($+0.0280$ to $+0.0984$) & $<0.001$ \\
CLAM-SB  & 0.6242 & $\mathbf{0.6414}$ & $+0.0172$~($-0.0025$ to $+0.0369$) & $0.037$  \\
TransMIL & 0.6627 & $\mathbf{0.7161}$ & $+0.0534$~($+0.0111$ to $+0.0957$) & $0.003$  \\
\bottomrule
\end{tabular}
\end{table}

\subsection{Bidirectional Cross-Cancer Consistency and Ablation Study}
\label{sec:res-bidirectional}
We evaluated CIT effectiveness in reverse-direction transfer (TCGA-STAD to TCGA-COAD) to assess generalization and robustness. Models trained on gastric tissue were evaluated zero-shot on colorectal cancer using the protocol in \cref{sec:res-transfer}. CIT improved MSI-H AUC for all three aggregators in reverse transfer as well (\cref{tab:reverse}). TransMIL showed $+0.041$ AUC, with smaller gains for ABMIL ($+0.016$) and CLAM-SB ($+0.006$). Improvements in both directions (COAD to STAD: $+0.0534$; STAD to COAD: $+0.0408$) suggest spatial immune topology is not cancer-type-specific and remains relatively stable under the distribution shifts tested.

Leave-one-group-out ablation (\cref{tab:ablation}) shows all descriptor groups contribute to cross-cancer transfer. Removal of immune--tumor mixing (G4) produced the largest decrease ($-0.0111$), followed by peritumoral margin (G2, $-0.0076$), TLS (G1, $-0.0046$), and multi-scale immune density (G3, $-0.0027$). Removing the two interface-oriented descriptors (G2 margin and G4 mixing) produced the two largest single-group drops (0.0187 AUC combined), indicating immune--tumor boundary interactions are important for cross-cancer transfer. 
\begin{table}[!tbp]
\caption{\small Reverse-direction zero-shot transfer, models trained on TCGA-STAD and tested on TCGA-COAD ($301$ patients; $54$ MSI-H / $247$ MSS). CIT improves MSI-H AUC across all aggregators, with the largest gain for TransMIL ($+0.041$).}
\label{tab:reverse}
\centering
\small
\setlength{\tabcolsep}{6pt}
\begin{tabular}{lccc}
\toprule
\textbf{Model} & \textbf{Baseline} & \textbf{CIT} & \textbf{$\Delta$} \\
\midrule
ABMIL    & $0.7066$ & $\mathbf{0.7228}$ & $+0.016$ \\
CLAM-SB  & $0.7490$ & $\mathbf{0.7547}$ & $+0.006$ \\
TransMIL & $0.7470$ & $\mathbf{0.7878}$ & $+0.041$ \\
\bottomrule
\end{tabular}
\end{table}
\begin{table}[!tbp]
\caption{\small Leave-one-group-out ablation (UNI2-h~$+$~TransMIL). Internal TCGA-COAD AUC (mean~$\pm$~SD) and zero-shot TCGA-STAD AUC with $\Delta$STAD relative to full CIT.}
\label{tab:ablation}
\centering
\small
\setlength{\tabcolsep}{6pt}
\begin{tabular}{lccc}
\toprule
\textbf{Configuration} & \textbf{COAD AUC} & \textbf{STAD AUC} & \textbf{$\Delta$STAD} \\
\midrule
Baseline (no CIT)          & $0.9398 \pm 0.0274$ & $0.6627$ & --- \\
\quad $-$~G1 (TLS)         & $0.9559 \pm 0.0115$ & $0.7115$ & $-0.0046$ \\
\quad $-$~G2 (margin)      & $0.9554 \pm 0.0117$ & $0.7085$ & $-0.0076$ \\
\quad $-$~G3 (TIL density) & $0.9562 \pm 0.0114$ & $0.7134$ & $-0.0027$ \\
\quad $-$~G4 (mixing)      & $0.9547 \pm 0.0119$ & $0.7050$ & $-0.0111$ \\
Full CIT (all 4 groups)    & $\mathbf{0.9567 \pm 0.0112}$ & $\mathbf{0.7161}$ & --- \\
\bottomrule
\end{tabular}
\end{table}
Ablations had negligible internal effect ($\Delta \leq 0.0020$ AUC), confirming CIT operates orthogonally to within-cohort discrimination; full representation is required for cross-cancer robustness. Removing any single descriptor group reduced zero-shot TCGA-STAD performance (\cref{tab:ablation}), indicating all four groups contribute to the total 0.0534 gain.
\section{Discussion}
\label{sec:discussion}
Foundation-model embeddings and CIT encode complementary signal: embeddings capture within-cohort appearance, while CIT captures cross-cohort immune organization. Foundation-model features degraded by 0.2771 AUC under cross-cancer transfer, whereas CIT-augmented features degraded by 0.2406 AUC, reducing the cross-cancer degradation by 0.0365 AUC. This pattern was consistent across all three MIL aggregators and both transfer directions. TransMIL showed asymmetric CIT gains: +0.0169 AUC internally (0.9398 to 0.9567, consistent with a ceiling effect) versus +0.0534 AUC cross-cancer (0.6627 to 0.7161, $p=0.003$), suggesting that CIT benefit is specific to conditions of distribution shift rather than uniform across settings. Although informative within a cohort, foundation-model embeddings tend to encode stain, scanner, and institution-specific variation rather than organ-invariant signal. CIT summarizes spatial immune organization motivated by MSI-associated immunobiology: TLS, peritumoral immune reactions, lymphocyte density, and immune--tumor mixing. These spatial descriptors appear less sensitive to scanner, institution, and tissue-type variation, which may explain the larger mean CIT gain under cross-cancer vs. cross-site transfer (0.0446 vs. 0.0253 AUC).

Ablation analysis demonstrates improvement does not depend on a single descriptor family. Removing any group reduced zero-shot TCGA-STAD performance; no individual group accounted for the full gain with complete representation. Unlike domain adaptation methods that require target-domain data~\cite{ref-cheung2026,ref-terashita2024}, CIT operates on frozen embeddings. In contrast to generic positional encodings that describe slide location, CIT couples spatial location to immune enrichment, encoding tissue organization without annotations~\cite{ref-corredor2019}. This augmentation increased zero-shot transfer AUC by 0.0534 using only ten additional descriptor dimensions per tile. Improvements across both transfer directions (STAD to COAD: $+0.0408$; COAD to STAD: $+0.0534$) suggest spatial immune topology may encode organ-invariant signal and remains relatively stable across the distribution shifts tested, indicating the effect is not specific to a single transfer direction.
\subsubsection{Limitations}
While CIT demonstrates promising generalization, the current study has two main limitations. First, our unsupervised immune clustering lacks explicit histological validation, necessitating future verification via pathologist annotations or cell-level segmentation. Second, because our evaluation is restricted to gastrointestinal cohorts (COAD and STAD), establishing broader pan-cancer generalizability requires further testing on non-gastrointestinal MSI-H datasets, such as TCGA-UCEC.

\section{Conclusion}
\label{sec:conclusion}
We present CIT, a lightweight spatial descriptor that captures conserved immune organization from frozen foundation-model embeddings without requiring dense annotations. By encoding key spatial immune patterns, CIT integrates seamlessly with existing MIL architectures. The proposed approach improved zero-shot cross-cancer MSI-H prediction across all aggregators while maintaining within-cohort accuracy, effectively mitigating the transfer degradation seen when using appearance-based features alone. These results suggest that spatial immune organization may serve as a fundamental, potentially organ-invariant biomarker. Future work will evaluate CIT on non-gastrointestinal datasets, such as endometrial and ovarian cancers, to establish its broader generalizability across diverse histopathological contexts.

\section*{Code Availability}
The implementation of CIT, pre-computed spatial descriptors, and evaluation scripts for the Conserved Immune Topology (CIT) framework are publicly available on GitHub at \url{https://github.com/raajuuu1998/cit_msih}.

\end{document}